%% file: vldb-main.tex
\PassOptionsToPackage{table}{xcolor}
\documentclass[sigconf, nonacm, pdfa]{acmart}

\newcommand\vldbdoi{10.14778/3827998.3828066}
\newcommand\vldbpages{4562 - 4565}
\newcommand\vldbvolume{19}
\newcommand\vldbissue{12}
\newcommand\vldbyear{2026}

\newcommand\vldbauthors{{Yifan Wu, Yuchen Peng, Jiaqi Chai, Yufei Qian, Xilin Li, Ke Chen, Lidan Shou}}
\newcommand\vldbtitle{Guixu: Valuation-Driven Data Discovery for Autonomous AI Agents with On-Chain Attestation}
\newcommand\vldbavailabilityurl{https://github.com/guixu-project/guixu}
\newcommand\vldbpagestyle{empty}

\input{packages}

\usepackage[utf8]{inputenc}
\usepackage[UKenglish]{babel}
\usepackage{colorprofiles}
\usepackage[a-2b]{pdfx}
\usepackage{fontawesome}
\usepackage[T1]{fontenc}
\usepackage{pifont}

\hypersetup{
  pdfstartview=,
  colorlinks=false,
  pdfborder={0 0 0},
  hidelinks,
  pdftitle={Guixu: Valuation-Driven Data Discovery for Autonomous AI Agents with On-Chain Attestation},
  pdfauthor={Yifan Wu},
  pdfsubject={Guixu: Valuation-Driven Data Discovery for Autonomous AI Agents with On-Chain Attestation},
  pdfkeywords={AI Agents, Data Discovery, Data Valuation, Blockchain}
}

\begin{document}

\title{\Guixu: Valuation-Driven Data Discovery for Autonomous AI Agents with On-Chain Attestation}

\input{main/authors}

\input{main/section/0_abstract}

\maketitle

\pagestyle{\vldbpagestyle}
\begingroup\small\noindent\raggedright\textbf{PVLDB Reference Format:}\\
\vldbauthors. \vldbtitle. PVLDB, \vldbvolume(\vldbissue): \vldbpages, \vldbyear.\\
doi:\vldbdoi
\endgroup
\begingroup
\renewcommand\thefootnote{}\footnote{\noindent
This work is licensed under the Creative Commons BY-NC-ND 4.0 International License. Visit \url{https://creativecommons.org/licenses/by-nc-nd/4.0/} to view a copy of this license. For any use beyond those covered by this license, obtain permission by emailing \href{mailto:info@vldb.org}{info@vldb.org}. Copyright is held by the owner/author(s). Publication rights licensed to the VLDB Endowment. \\
\raggedright Proceedings of the VLDB Endowment, Vol. \vldbvolume, No. \vldbissue\%
ISSN 2150-8097. \\
\href{https://doi.org/\vldbdoi}{doi:\vldbdoi} \\
}\addtocounter{footnote}{-1}\endgroup

\ifdefempty{\vldbavailabilityurl}{}{
\vspace{.3cm}
\begingroup\small\noindent\raggedright\textbf{PVLDB Artifact Availability:}\\
The source code, data, and/or other artifacts have been made available at \url{\vldbavailabilityurl}.
\endgroup
}

\input{main/section/1_introduction}
\input{main/section/2_overview}

\input{main/section/3_demonstration}

\begin{acks}
This work was supported by the Pioneer R\&D Program of Zhejiang (No. 2024C01021) and the Leading Talent of Technological Innovation Program (No. 2023R5214) of Zhejiang Province.
\end{acks}

\bibliographystyle{ACM-Reference-Format}
\bibliography{main/references}

\end{document}

%% file: packages.tex
\usepackage{amsmath}
\usepackage{graphicx}
\usepackage{textcomp}
\usepackage{xcolor}
\usepackage[normalem]{ulem}
\usepackage{listings}
\usepackage[linesnumbered,boxed,ruled,vlined,commentsnumbered]{algorithm2e}
\usepackage{courier}
\usepackage{booktabs}
\usepackage{color}
\usepackage{multirow}
\usepackage{makecell}
\usepackage{wrapfig}
\usepackage{url}
\usepackage{ifthen}
\usepackage{balance}
\usepackage{forest}
\usepackage{pifont}
\usepackage{adjustbox}
\usepackage[shortlabels]{enumitem}
\usepackage{soul}
\usepackage{xcolor}
\usepackage{multicol}
\usepackage{xparse}
\usepackage{mdframed}
\usepackage{bookmark}
\usepackage[utf8]{inputenc}

\usepackage{arydshln}
\usepackage{amsthm}

\SetKwProg{Fn}{Function}{}{end}
\SetKwProg{Proc}{Procedure}{}{end}

\definecolor{mymauve}{rgb}{0.58,0,0.82}
\definecolor{dkgreen}{rgb}{0,0.6,0}
\definecolor{browncolor}{rgb}{0.6, 0.3, 0.0}

\definecolor{lowdpcolor}{rgb}{0.9, 0.8, 0.2}
\definecolor{lightpurple}{rgb}{0.8, 0.7, 0.9}
\definecolor{lightpink}{rgb}{1.0, 0.8, 0.86}

\definecolor{advantagecolor}{rgb}{0.8, 1.0, 0.8}%
\definecolor{disadvantagecolor}{HTML}{FFE5E5}%

\definecolor{lowdpcolor}{HTML}{DDDDFF}
\definecolor{highdpcolor}{HTML}{FFCCCC}

\definecolor{lightred}{HTML}{FFCCCC}%
\definecolor{lightblue}{HTML}{CCCCFF}%

\setenumerate[1]{itemsep=0pt,partopsep=0pt,parsep=\parskip,topsep=0.5pt}
\setitemize[1]{itemsep=0pt,partopsep=0pt,parsep=\parskip,topsep=0.5pt}
\setdescription{itemsep=0pt,partopsep=0pt,parsep=\parskip,topsep=0.5pt}

\definecolor{myblue}{rgb}{0,0,1}
\definecolor{verylightblue}{rgb}{0.8, 0.9, 1.0}
\sethlcolor{verylightblue}

\forestset{
  default preamble={
    for tree={
      parent anchor=south,
      child anchor=north,
      align=center,
      edge={-latex},
      rounded corners,
      draw,
      fill=white,
      s sep=7mm,
      l sep=10mm,
      anchor=center,
      calign=center,
      align=center
    }
  }
}

\usepackage{caption}
\def\BibTeX{{\rm B\kern-.05em{\sc i\kern-.025em b}\kern-.08em
    T\kern-.1667em\lower.7ex\hbox{E}\kern-.125emX}}

\newtheorem*{scenario*}{\bf Targeted Scenarios}

\newcommand{\Guixu}{\textsc{Guixu}\xspace}

\useunder{\uline}{\ul}{}

\usepackage[skins,breakable]{tcolorbox}

\renewcommand{\shortauthors}{Yifan Wu et al.}

%% file: main/authors.tex
\settopmatter{authorsperrow=4}
\author{Yifan Wu}
\authornote{These authors contributed equally to this work.}
\affiliation{%
  \institution{Zhejiang University}
}
\email{yifan.wu@zju.edu.cn}

\author{Yuchen Peng}
\authornotemark[1]
\affiliation{%
  \institution{Zhejiang University}
}
\email{zjupengyc@zju.edu.cn}

\author{Jiaqi Chai}
\authornotemark[1]
\affiliation{%
  \institution{Zhejiang University}
}
\email{chaijiaqi@zju.edu.cn}

\author{Yufei Qian}
\authornotemark[1]
\affiliation{%
  \institution{Zhejiang University}
}
\email{qianyufei@zju.edu.cn}

\author{Xilin Li}
\affiliation{%
  \institution{Zhejiang University}
}
\email{lixilin@zju.edu.cn}

\author{Ke Chen}
\affiliation{%
  \institution{Zhejiang University}
}
\email{chenk@zju.edu.cn}

\author{Lidan Shou}
\authornote{Lidan Shou is the corresponding author.}
\affiliation{%
  \institution{Zhejiang University\authornote{Also affiliated with: The State Key Laboratory of Blockchain and Data Security; Hangzhou High-Tech Zone (Binjiang) Institute of Blockchain and Data Security}}
}
\email{should@zju.edu.cn}

%% file: main/section/0_abstract.tex
\begin{abstract}
Autonomous agents increasingly rely on external data to complete downstream tasks such as model training and decision support. However, existing data discovery systems remain largely retrieval-oriented: they surface candidate datasets from heterogeneous sources, but provide limited support for estimating task-specific utility, selecting cost-effective datasets under budget constraints, or incorporating trustworthy feedback from prior usage. This paper presents \Guixu, a valuation-driven data discovery system for autonomous agents. \Guixu employs a three-phase valuation pipeline with proxy-label propagation and multi-round knapsack optimization for task-aware data valuation. \Guixu integrates agentic payment protocol to enable budget-constrained data procurement workflows.
\Guixu leverages on-chain data market and attestation signals for verifiable data discovery. Our demonstration highlights how \Guixu enables an agent to move beyond keyword-based dataset retrieval toward task- and budget-aware, trustworthy data discovery and procurement. Attendees can interactively explore the full workflow, from NL task specification and multi-source search to data valuation and verifiable transaction feedback.
\end{abstract}

%% file: main/section/1_introduction.tex
\section{Introduction} \label{section:introduction}
\begin{figure}[t!]
	 \centering
  \includegraphics[width=0.47\textwidth]{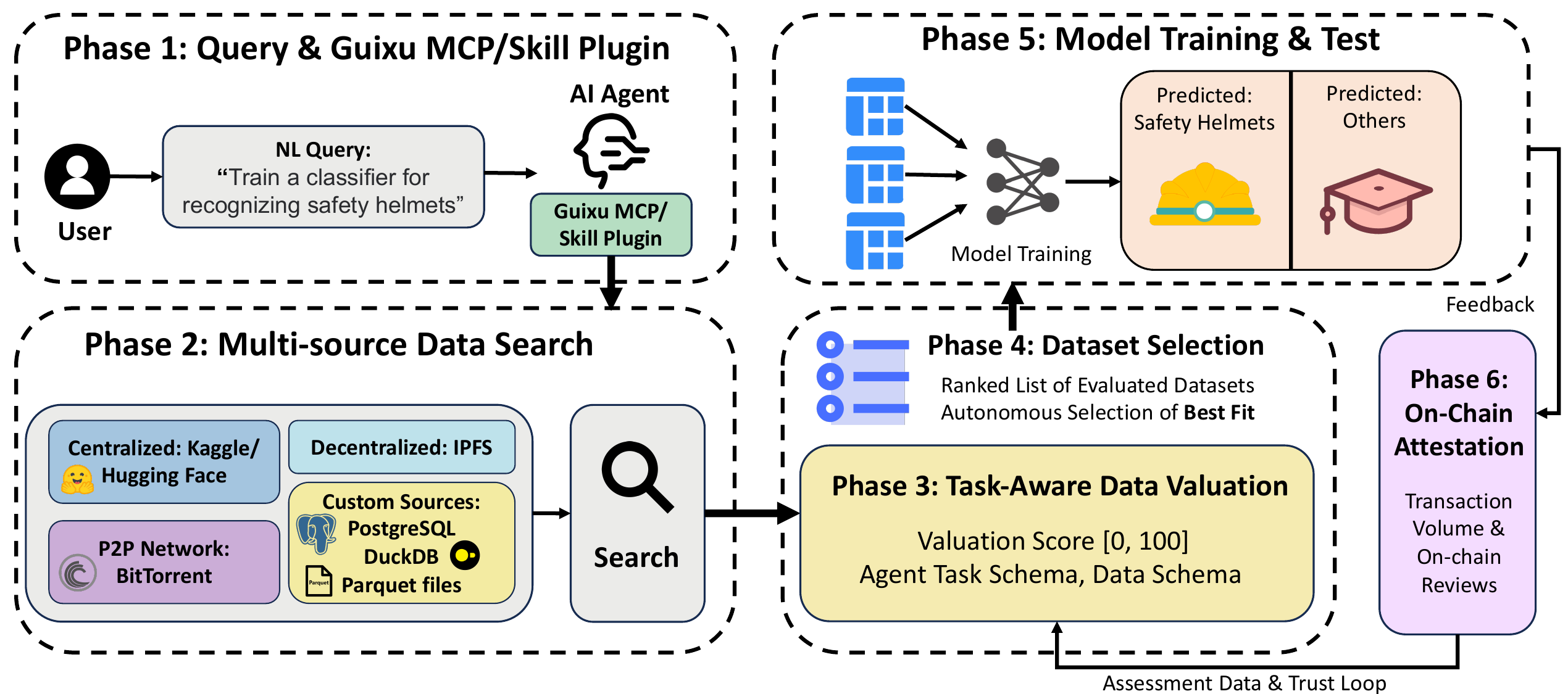}
	 \caption{Overview of the \Guixu workflow. A user's natural-language query is processed through six phases.}
     \label{figure:Guixu_role}
\end{figure}

Data discovery, defined as the process of identifying datasets that satisfy the information needs of data-driven applications~\cite{paton2023survey}, has become a foundational capability in modern AI and data systems. Its importance is amplified in agentic settings, where autonomous agents such as OpenClaw and OpenAI Codex must independently locate, evaluate, and procure external data to complete complex tasks, including model training, prediction, and decision support. In such workflows, data discovery is not merely a retrieval problem: agents must determine which datasets are genuinely useful for a given objective, which combination is worth acquiring under a budget constraint, and which sources are trustworthy enough to use without human oversight.

\textbf{State of the Art and Limitations.}
Existing dataset search platforms, including Google Dataset Search~\cite{Brickley2019GoogleDataset}, Kaggle, and Hugging Face, have exposed agent-accessible interfaces via Model Context Protocol (MCP) or skill plugins, enabling keyword-based search across heterogeneous sources. However, these systems operate in a fundamentally \textit{retrieval-first} manner: they surface candidate datasets but provide no support for task-aware data valuation, budget-constrained selection, or trust-aware procurement. Autonomous agents, therefore, still lack an integrated mechanism for translating a high-level task request into a cost-effective data acquisition decision.

To make this gap concrete, consider the following running example used throughout this paper: \textit{``Train an image classifier that determines whether workers in construction site images are wearing safety helmets correctly, with a total data procurement budget of \$2.00.''} A practical system must do far more than retrieve datasets containing the keyword \textit{helmet}: it must quantify each dataset's utility relative to the task objective, account for acquisition cost, and incorporate trust signals when deciding whether a source is worth purchasing. Existing tools leave these decisions to manual inspection, which is fundamentally misaligned with autonomous agent execution.

Specifically, we identify three limitations of existing data discovery systems. \textbf{(1) Insufficient Task-Aware Data Valuation:} Existing systems lack a standardized schema of agent tasks and their associated data requirements, and thus cannot reason about which datasets are genuinely valuable for a given task beyond superficial keyword relevance. This causes agents to surface functionally incompatible datasets, such as massive image sets that trigger processing timeouts due to task-irrelevant samples, or datasets whose volume fails to meet the minimum execution thresholds of agent-generated code.
\textbf{(2) Incompatibility with Agentic Payment:} Existing platforms do not support decentralized agentic payment protocols such as x402~\cite{x402} and machine payment protocol (MPP)~\cite{mpp}. Datasets relevant to niche scenarios are rarely free, yet how to optimally procure a portfolio of datasets under a fixed budget remains largely unaddressed.
\textbf{(3) Absence of On-Chain Attestation Signals:} Blockchain ledgers expose valuable trust signals (e.g., transaction volume, on-chain reviews, and seller reputation) that can substantially improve data valuation, yet existing discovery systems rarely incorporate them.

\textbf{Our Approach.}
This paper presents \Guixu, a valuation-driven data discovery system for autonomous agents. \Guixu reframes dataset discovery as a \emph{task-aware valuation} and \emph{procurement} problem rather than a pure retrieval problem.
As illustrated in Figure~\ref{figure:Guixu_role}, a user's natural-language query flows through six phases: NL query parsing, multi-source data search, task-aware valuation, autonomous dataset selection, model training and testing, and on-chain attestation that feeds trust signals back into future valuations. This end-to-end workflow is realized by three tightly integrated layers.

First, the \emph{frontend and protocol layer} provides the \Guixu Hub marketplace and an MCP server interface through which AI agents access all discovery capabilities, together with a multi-protocol agentic payment framework that dynamically routes transactions via the x402 agentic payment protocol or an on-chain escrow contract depending on the transaction context.
Second, the \emph{valuation-driven discovery layer} parses an NL query into a structured task schema and data schema, searches heterogeneous data sources, and scores each candidate dataset through a multi-signal coarse ranking, a proxy-label propagation phase, and a multi-round knapsack optimization, thereby enabling agents to autonomously select the most valuable datasets for a given task.
Third, the \emph{on-chain data market layer} handles trustworthy data transactions via escrow-based smart contracts and a decentralized key management network, while its on-chain attestation module records buyer reviews and seller reputation and feeds these trust signals back into the valuation model to continuously improve discovery quality.
In our running example, \Guixu enables an agent to autonomously discover, evaluate, purchase, and assemble helmet-detection training data, completing the loop from natural-language intent to verified model training.
We view \Guixu as an initial system step toward more autonomous and trustworthy data discovery and procurement workflows.

\textbf{Audience Interaction}. \Guixu is designed as a live, interactive system demonstration. Users can submit natural-language tasks, inspect candidate datasets collected from multiple sources, observe how valuation signals affect ranking and dataset portfolio selection, and complete a verifiable on-blockchain procurement workflow. The system therefore demonstrates not only how autonomous agents can search for data, but also how they can reason about data utility, cost, and trust in a principled and operational manner.

%% file: main/section/2_overview.tex
\section{Overview of \Guixu} \label{section:overview}
\Guixu comprises three layers: a \emph{frontend and protocol layer} (Section~\ref{subsection:guixu-frontend-mcp}) exposing the Hub marketplace, MCP interface, and agentic payment framework; a \emph{valuation-driven discovery layer} (Section~\ref{subsection:data-valuation}) transforming natural-language queries into optimal dataset portfolios; and an \emph{on-chain data market layer} (Section~\ref{subsection:on-chain-market}) handling trustworthy transactions via escrow contracts, decentralized key management, and on-chain attestation.

\subsection{\Guixu Frontend and Protocols}\label{subsection:guixu-frontend-mcp}
Next, we delineate the key components of the \Guixu ecosystem: the \Guixu Hub frontend module, the agent-native MCP plugin, and a specialized protocol tailored for autonomous agentic payment.

\noindent\textbf{\Guixu Hub Platform.} \Guixu Hub serves as the marketplace frontend, providing cataloging, data discovery, and transaction management for data assets. Sellers publish datasets with structured metadata, including schema definitions, semantic tags, pricing, and access policies. Consumers interact via full-text search, on-chain transaction synchronization, and on-chain review aggregation.

\noindent\textbf{Model Context Protocol.} \Guixu encapsulates all data acquisition capabilities as a Model Context Protocol (MCP) server, exposing them to AI agents via JSON-RPC over stdio and HTTP transports. The server advertises a typed tool catalog spanning intent parsing, dataset search, quality evaluation, purchase, and feedback, enabling an agent to orchestrate the complete data procurement pipeline within a single MCP session.

\noindent\textbf{Agentic Payment Protocol.} \Guixu supports a multi-protocol agentic payment framework (e.g., x402 and MPP) for autonomous agent transactions. A payment router inspects the transaction context to dynamically select the settlement mechanism: micropayments default to the x402 protocol with atomic on-chain settlement, whereas high-value transactions are routed to an escrow contract that supports fund locking.

\subsection{Valuation-Driven Data Discovery}\label{subsection:data-valuation}

Once the agent has connected to \Guixu via the MCP interface and configured its payment credentials through the protocol layer above, the system proceeds to the core discovery pipeline.

\noindent\textbf{NL Query Parser.} A natural-language intent parser transforms the user's free-form query into a structured profile guiding all downstream stages. It issues a zero-temperature, JSON-constrained LLM call to extract a task schema (e.g., task type, target entity, keywords) and a data schema (e.g., budget ceiling, dataset size bounds, column names). The profile is further enriched with local context, such as hardware profiles and agent memory, ensuring it reflects both explicit intent and implicit computational constraints.

\begin{figure}[t!]
	 \centering
  \includegraphics[width=0.42\textwidth]{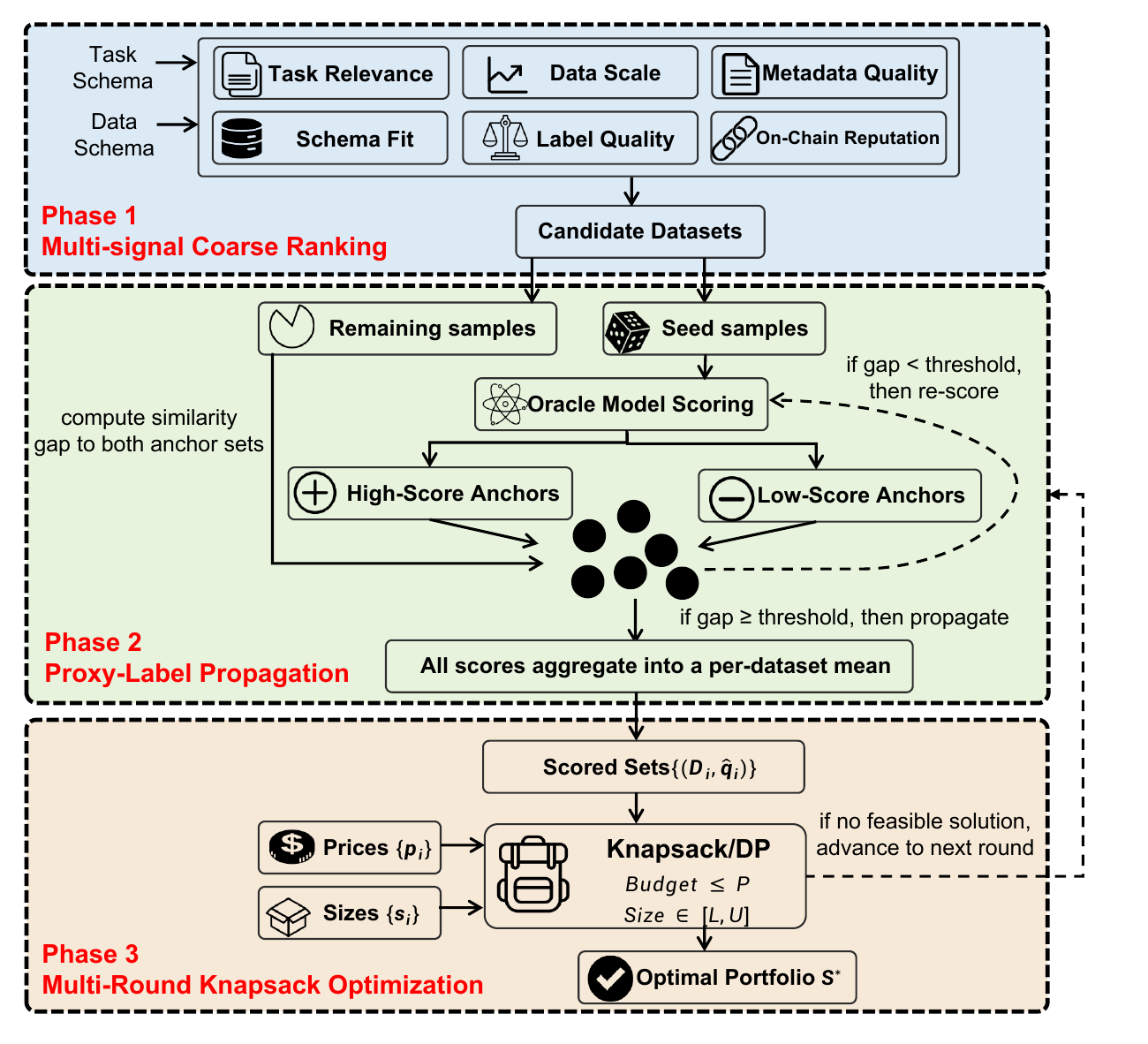}
	 \caption{Three-phase pipeline of task-aware data valuation.}
     \label{figure:data_valuation_figure}
\end{figure}

\noindent\textbf{Task-aware Data Valuation.} Given the parsed query profile and the candidate datasets returned by the \Guixu multi-source search module, the valuation module estimates the task-specific utility of each dataset through a three-phase pipeline, as shown in Figure~\ref{figure:data_valuation_figure}.

\emph{Phase~1: Multi-Signal Coarse Ranking.}
Each candidate dataset is scored by a weighted combination of five task-aware signals: (i)~\emph{relevance}, cosine similarity between the task description embedding and dataset representation; (ii)~\emph{schema fit}, the degree to which the dataset schema satisfies task-required columns; (iii)~\emph{data scale}, a saturating logarithmic function of sample count and byte size; (iv)~\emph{label quality}, a composite indicator of class balance, missing-value rate, and target-column annotations; and (v)~\emph{metadata completeness}, measuring the availability of descriptions, summaries, and provenance metadata.
(vi)~\emph{on-chain reputation}, an additive adjustment blending trade activity frequency with aggregated review sentiment from the attestation layer (Section~\ref{subsection:on-chain-market}).
The top-$k$ datasets are selected to proceed as a batch.
\Guixu computes the heuristic weight of each signal using the Shapley value algorithm, based on analysis of historical dataset performance.

\emph{Phase~2: Proxy-Label Propagation.}
To estimate per-sample utility without invoking an expensive LLM oracle on every sample, the system employs a proxy-label propagation strategy. For each candidate dataset, a small seed subset of samples is randomly selected and scored by a strong generative model (the \emph{LLM oracle}), producing per-sample utility judgments. The scored seeds are then partitioned into \emph{high-score anchors} and \emph{low-score anchors} based on configurable thresholds. For each remaining unscored sample, the system computes its average textual similarity to both anchor sets. If the similarity gap $\Delta$ between the two anchor affinities exceeds a user-configurable threshold~$\theta$, the sample inherits the mean score of its nearest anchor set. Larger $\theta$ means higher accuracy but more oracle cost, and vice versa.
Samples for which $\Delta < \theta$ are deemed \emph{uncertain} and are sent back to the LLM oracle for re-scoring. The final per-dataset quality score~$\hat{q}_i$ is the mean of all sample-level scores (seed, propagated, and re-judged). This design ensures that the majority of samples are scored via lightweight similarity propagation, while LLM invocations are reserved for the small fraction of ambiguous cases, yielding substantial cost savings without sacrificing scoring fidelity.

\emph{Phase~3: Multi-Round Knapsack Optimization.}
Given the scored candidates $\{(D_i, \hat{q}_i)\}$ together with their prices $\{p_i\}$ and sizes $\{s_i\}$, the system selects an optimal dataset portfolio $S^*$ that maximizes aggregate utility subject to a budget constraint $\sum_{i \in S} p_i \le P$ and a size constraint $L \le \sum_{i \in S} s_i \le U$. This is formulated as a two-dimensional knapsack problem and solved via dynamic programming over discretized price units, with a size-indexed frontier at each budget level to prune dominated states. If no feasible solution exists within the current candidate batch, the system expands the pool by advancing to the next batch of coarse-ranked datasets, triggering a new round of sample-efficient scoring before re-running the optimization; hence the term \emph{multi-round}. The process terminates when a feasible portfolio is found or all batches are exhausted.

\subsection{On-Chain Data Market}\label{subsection:on-chain-market}

\begin{figure}[t!]
	 \centering
  \includegraphics[width=0.40\textwidth]{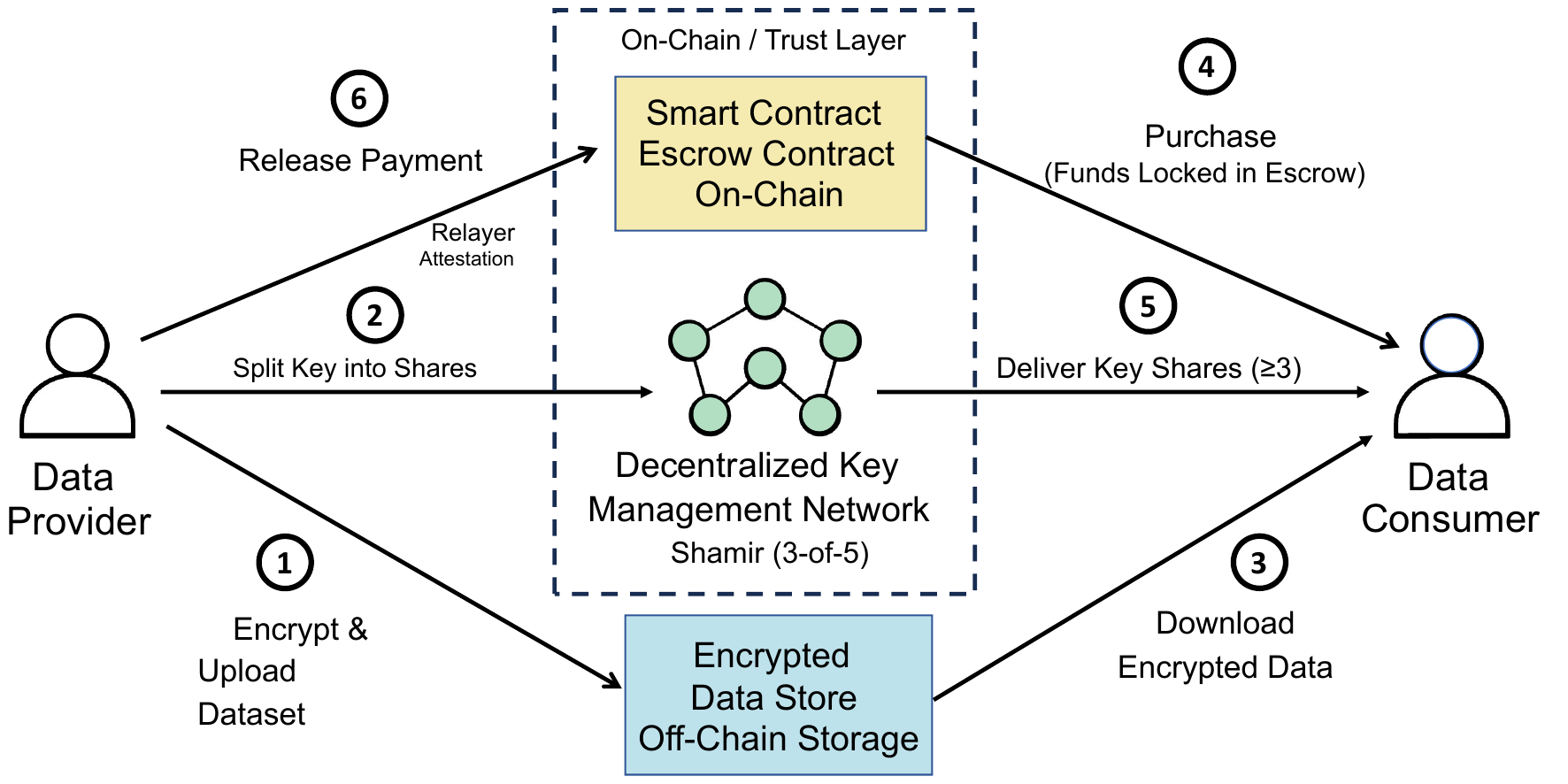}
	 \caption{End-to-end transaction lifecycle of the on-chain data market.}
     \label{figure:on_chain_market}
\end{figure}

\textbf{Overview}.
As illustrated in Figure~\ref{figure:on_chain_market}, the end-to-end transaction lifecycle proceeds in six stages: \ding{182}~the provider encrypts the dataset and uploads the encrypted datasets to off-chain storage; \ding{183}~the decryption key is split into five shares via Shamir's Secret Sharing and distributed to independent Keyguard nodes in the \Guixu Decentralized Key Management Network; \ding{184}~the data consumer locates the encrypted file by its content identifier and downloads it from off-chain storage; \ding{185}~the consumer initiates an on-chain purchase, locking funds in the listing's escrow contract; \ding{186}~a designated relayer verifies successful key-share delivery by collecting delivery receipts and submits an attestation transaction that triggers escrow release; \ding{187}~upon payment confirmation, at least three Keyguard nodes deliver their shares to the consumer, who reconstructs the decryption key locally.

\noindent\textbf{On-Chain Attestation.} Each dataset is governed by a smart contract deployed on the \emph{Base} chain that encodes the listing price, payment token, and cryptographic commitments. Upon purchase, funds are locked in escrow and released only upon confirmed delivery, with a refund option after a protocol-defined timeout. \Guixu derives trust signals directly from on-chain evidence: a chain indexing client extracts consumer reviews from purchase transaction metadata and aggregates them into per-listing summaries that feed into the valuation module's on-chain signal dimension. In parallel, a seller reputation module computes a profile based on sales volume, consumer count, and activity span, mapped to a tiered trust classification. Each dataset is further accompanied by a W3C Verifiable Credential attesting to its content-addressed identifier, providing a cryptographically verifiable provenance chain.

\noindent\textbf{Decentralized Key Management Network (DKMN).} \Guixu Hub is the data market to apply decentralized key management techniques to dataset key custody, instantiating a 3-of-5 Shamir Secret Sharing scheme across independent Keyguard nodes via the Lit Protocol~\cite{lit-protocol} to guarantee that no single party ever holds the complete decryption key.

%% file: main/section/3_demonstration.tex
\begin{figure*}[!t]
    \centering
    \includegraphics[width=0.98\linewidth]{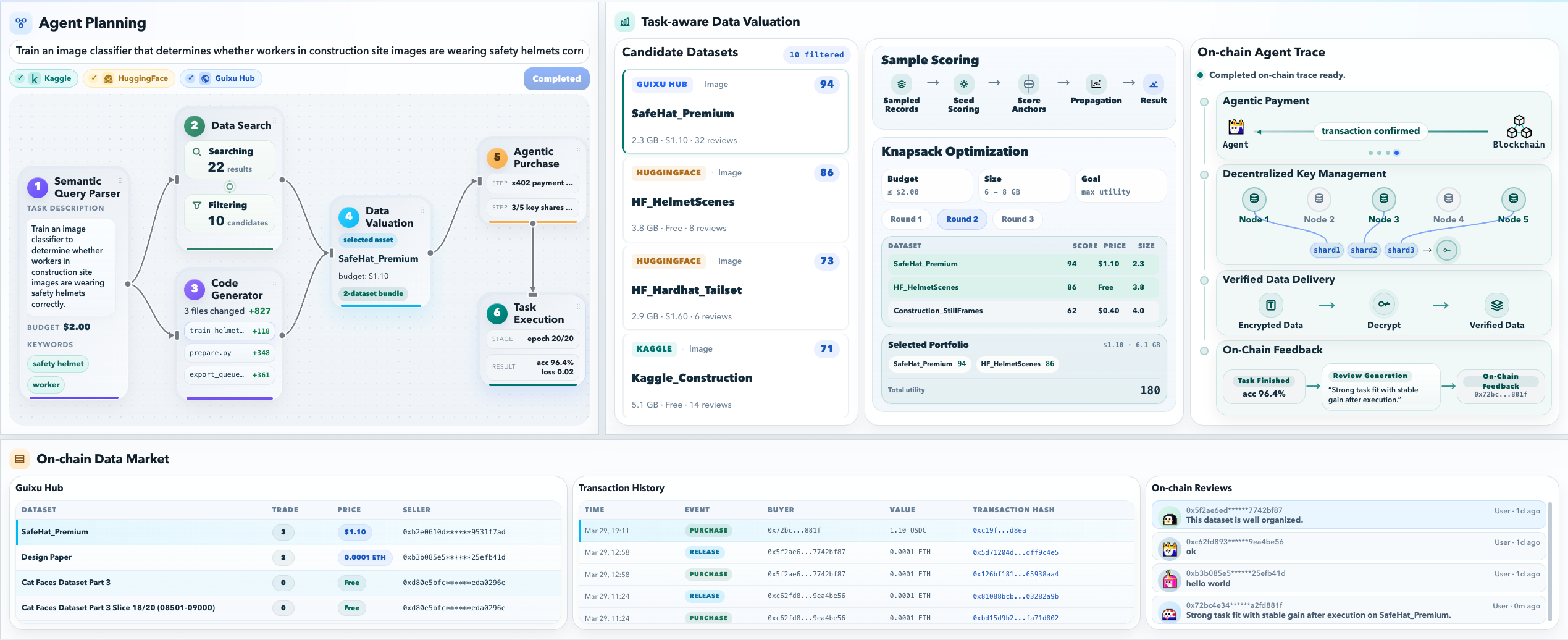}
    \caption{The Graphical User Interface of \Guixu.}
    \label{fig:demo_interface}
\end{figure*}

\section{Demonstration} \label{section:demonstration}
\textbf{Demonstration Setup and Interaction.}
\Guixu showcases an agent-native interface for valuation-driven data discovery and data acquisition for autonomous AI agents.
The interface is connected to the \Guixu MCP server, the \Guixu Hub backend, heterogeneous data sources, and smart contracts.
Attendees will be able to observe how \Guixu turns a natural-language task into agentic planning, task-aware data valuation, and on-chain interaction under budget constraints.
As shown in Figure~\ref{fig:demo_interface}, we provide two demonstration scenarios by varying the agent's task and budget inputs.

\noindent\textbf{Scenario 1: Data Discovery with No Budget.}
\emph{Our first demonstration evaluates how agents rely only on free datasets to build classification models.}
The Agent first issues a task to check whether the cat is captured by the monitor with no budget.
In response, the \emph{Agent Planning} panel shows how \Guixu parses the request into a structured task profile and automatically restricts discovery to free sources such as Kaggle and Hugging Face.
The \emph{Task-aware Data Valuation} panel then visualizes the returned candidate datasets, the sample-scoring process, and the knapsack-based selector under the zero-budget constraint, allowing attendees to inspect how data is chosen to maximize task utility.
Throughout this scenario, attendees can click any operator nodes and dataset panels to inspect intermediate outputs and observe that the valuation-driven selection yields a measurable accuracy gain even at zero budget.

\noindent\textbf{Scenario 2: Data Acquisition under Budget Constraints.}
\emph{Our second demonstration evaluates how agents autonomously select paid datasets when higher task utility justifies the cost.}
The agent starts the safety-helmet classification task, but with a budget of \$2.00 and access to the \Guixu Hub.
The \emph{Agent Planning} and \emph{Task-aware Data Valuation} panels indicate that paid datasets are involved, and the optimization process also incorporates budget constraints to maximize utility.
In the \emph{On-chain Agent Trace} panel, attendees can follow the full transaction life cycle: agentic payment, decentralized key management, and verified data delivery.
In parallel, the \emph{On-Chain Data Market} panel displays the dataset listing, its full transaction history, and accumulated on-chain reviews.

\noindent\textbf{Key Benefits for the Audience.}
\Guixu offers attendees three distinctive perspectives on agent-native data discovery:
\begin{itemize}[leftmargin=1em,itemsep=0em,topsep=0em]
    \item \textbf{Agent-Native Data Discovery.} Unlike conventional dataset search that matches keywords against metadata, \Guixu lets attendees experience a discovery paradigm driven by an autonomous agent that reasons about task requirements and market feedback. This gives the audience deep insights into how task-defined utility reshapes dataset selection.

    \item \textbf{Transparent and Interpretable Valuation.} \Guixu visualizes each stage of its valuation pipeline, from coarse ranking signals and sample-level scoring to knapsack optimization. This allows attendees to inspect not only \textit{which} datasets are selected, but also \textit{why} and how the selection is made.

    \item \textbf{Interactive On-Chain Attestation.}
    \Guixu further enables attendees to inspect how on-chain attestation transforms each transaction into a verifiable, traceable event. By examining transaction history and on-chain reviews, the audience can assess the market's trustworthiness and how these attestation signals are incorporated into future valuation decisions.
\end{itemize}

%% file: main/references.bib
@String{Computing = "Computing" }

@misc{x402,
  author = {Coinbase},
  title = {x402},
  year = {2026},
  howpublished = {\url{https://www.x402.org/}},
  note = {Accessed: 2026-03-29}
}

@misc{mpp,
  author = {Stripe},
  title = {Machine Payment Protocol},
  year = {2026},
  howpublished = {\url{https://mpp.dev/}},
  note = {Accessed: 2026-03-29}
}

@misc{lit-protocol,
  author = {Lit Protocol},
  title = {Lit Protocol},
  year = {2026},
  howpublished = {\url{https://litprotocol.com}},
  note = {Accessed: 2026-03-29}
}

@misc{Kaggle,
  author = {Kaggle},
  title = {Kaggle},
  year = {2026},
  howpublished = {\url{https://www.Kaggle.com}},
  note = {Accessed: 2026-03-29}
}

@inproceedings{Brickley2019GoogleDataset,
author = {Brickley, Dan and Burgess, Matthew and Noy, Natasha},
title = {Google Dataset Search: Building a search engine for datasets in an open Web ecosystem},
year = {2019},
isbn = {9781450366748},
publisher = {Association for Computing Machinery},
address = {New York, NY, USA},
booktitle = {The World Wide Web Conference},
pages = {1365–1375},
numpages = {11},
location = {San Francisco, CA, USA},
series = {WWW '19}
}

@article{paton2023survey,
author = {Paton, Norman W. and Chen, Jiaoyan and Wu, Zhenyu},
title = {Dataset Discovery and Exploration: A Survey},
year = {2023},
issue_date = {April 2024},
publisher = {Association for Computing Machinery},
address = {New York, NY, USA},
volume = {56},
number = {4},
issn = {0360-0300},
journal = {ACM Comput. Surv.},
month = nov,
articleno = {102},
numpages = {37}
}
